# Imitation Learning with a Value-Based Prior

**Umar Syed**
Computer Science Department
Princeton University
35 Olden St.
Princeton, NJ 08540-5233
usyed@cs.princeton.edu

**Robert E. Schapire**
Computer Science Department
Princeton University
35 Olden St.
Princeton, NJ 08540-5233
schapire@cs.princeton.edu

## Abstract

The goal of *imitation learning* is for an apprentice to learn how to behave in a stochastic environment by observing a mentor demonstrating the correct behavior. Accurate prior knowledge about the correct behavior can reduce the need for demonstrations from the mentor. We present a novel approach to encoding prior knowledge about the correct behavior, where we assume that this prior knowledge takes the form of a Markov Decision Process (MDP) that is used by the apprentice as a rough and imperfect model of the mentor's behavior. Specifically, taking a Bayesian approach, we treat the value of a policy in this modeling MDP as the log prior probability of the policy. In other words, we assume *a priori* that the mentor's behavior is likely to be a high-value policy in the modeling MDP, though quite possibly different from the optimal policy. We describe an efficient algorithm that, given a modeling MDP and a set of demonstrations by a mentor, provably converges to a stationary point of the log posterior of the mentor's policy, where the posterior is computed with respect to the "value-based" prior. We also present empirical evidence that this prior does in fact speed learning of the mentor's policy, and is an improvement in our experiments over similar previous methods.

## 1 Introduction

Imitation learning and reinforcement learning can be viewed as two approaches to solving the same problem: learning how to behave in a stochastic environment. In each, the goal is to learn the best *policy*, i.e., a function mapping each of the environment's possible states to a distribution over actions that can be taken in that state. The two approaches differ in how they define the "best" policy, and in what they assume is available to a learning algorithm. In imitation learning, one assumes an apprentice has access to a set of examples (trajectories of state-action pairs) from a mentor's policy, which is also defined to be the best policy. Imitation learning can therefore be suc-cinctly described as "supervised learning of behavior". In reinforcement learning, one instead assumes the existence of a *reward function*, i.e., a mapping from each of the environment's states to a numerical reward. The best policy is defined to be the one that maximizes expected cumulative (and possibly discounted) reward.

Each of these approaches has its drawbacks. In imitation learning, as in any supervised learning problem, data from the mentor will typically be limited, particularly if the state space is large. In such cases, incorporating prior knowledge about the best policy (sometimes called *regularization*) can effectively compensate for a lack of data.

Reinforcement learning suffers from a more subtle (and usually unmentioned) disadvantage: it requires a way to access the true reward function. In principle, the reward function is provided by "nature", and is specified as part of the problem description. One either assumes that the rewards are available to the learning algorithm in explicit functional form, or assumes that they can be estimated from experience. In practice, however, rewards are usually specified by hand, and often need to be tweaked and tuned to elicit the desired behavior. Whenever this happens, it is misleading to treat the reward function as necessarily correct.

In this paper, we take a middle approach based on a *value-based prior*. We define the best policy to be the mentor's policy, and we use a *modeling MDP* to encode the apprentice's prior belief about the mentor's policy. We assume that the prior probability of any policy being the mentor's increases with the value of that policy in the modeling MDP. In this way, instead of relying solely on rewards or soley on evidence, the apprentice smoothly integrates both prior knowledge and observed information about the best policy.

For examples of when this may be a good idea, consider the problem of *dialog management*, the motivating application for our work. A dialog manager is a program that controls the actions of an automated telephone agent, such as the kind one encounters when calling a company's customer service number. Instead of asking the caller to navigate menus by pressing buttons, these agents encourage customers to speak freely, and attempt to offer an experience comparable to that of speaking to a live operator. The dialog manager makes decisions about which questions to ask, how to deal with unexpected responses, what to do when the customer is misunderstood (ask them to clarify? make a best guess and move on?), and when to give up and transfer the customer to an actual person.

There has been success in training dialog managers from data using reinforcement learning [6, 10]. However, this approach requires the assertion of a reward function that is based largely on intuition, since customers rarely give a clear indication about whether they are satisfied with a dialog. Indeed, Walker *et al* [12] have shown that evaluating the performance of a dialog manager is itself a challenging task, which calls into question whether reinforcement learning is sufficient to solve this problem, and suggests that some form of imitation may be needed.

Another challenge is the scarcity of suitable training opportunities. Observe that new dialog management strategies cannot be tested on a static corpus. They have to be tried in real dialogs with actual users, which is, needless to say, an expensive proposition. As a result, there has been much interest in building user models, i.e., simulators that mimic the behavior of customers. Schatzman *et al* [9] provide a survey and comparison of some attempts at learning user models from data. A common theme in recent work has been to leverage prior knowledge, and restrict the space of models to those that encode realistic user behavior, in the hope that less data will be needed for training.

The work in this paper has been developed with these issues in mind. At the same time, the framework and algorithms presented here are intended to be completely general, and not specific to dialog management. We assume that an apprentice is observing a mentor acting in a stochastic environment, and that the apprentice wants to estimate a model of the mentor's behavior. We furthur assume that the mentor is behaving in a roughly reward-seeking manner. The apprentice uses the value function of a modeling MDP to help guide its estimate towards the correct policy. For example, in the domain of dialog management, we can assign higher rewards in the modeling MDP to states that are closer to the end of the conversation. In this way, we can leverage our knowledge that customers and operators are both trying to complete their conversations as soon as possible, without needing to specify exactly how they are trying to accomplish that goal.

The paper proceeds as follows. After reviewing related work, we propose a formal definition of a prior distribution for the mentor's policy based on the value function of the modeling MDP. We next give our main theoretical contribution of this paper, which is an efficient algorithm for finding a stationary point of the log posterior distribution that is computed with respect to our novel prior. Finally we present experimental evidence, which we use to compare to previous methods, indicating that a value-based prior does speed the estimation of the mentor's policy.

## 2   Related Work

A number of authors have suggested methods to incorporate prior knowledge of the mentor's behavior into imitation learning. Price *et al* [7] described an approach based on the Dirichlet distribution. Henderson *et al* [4] developed a modified temporal difference learning algorithm in which the usual $Q$ values are adjusted so that the resulting optimal policy is forced to more closely match the mentor's behavior. Very recently, Fern *et al* [3] proposed a similar yet simpler method that uses a Boltzmann distribution to assign greater prior probability to mentor actions that have higher $Q$ values. In Section 5, we will empirically compare the methods of Price *et al* and Henderson *et al* to our algorithm.

Two recent papers by Abbeel *et al* [1] and Ratliff *et al* [8] have used *inverse reinforcement learning* (IRL) as a way to extract information from a mentor's demonstrations. In IRL, we are given a policy, or demonstrations from a policy, and the goal is find a reward function for which that policy is (near) optimal. Ratliff *et al* introduced a variant that favors those reward functions for which the optimal policy is similar to the observed policy, making their algorithm a type of imitation learning. Both papers assumed that the true reward function can be expressed as a linear combination of a set of known features, and leverage this assumption. Our work, by contrast, allows for arbitrary rewards, which we assume are given, but uses those rewards only to bias the inference of the mentor's policy.

## 3   Problem Formulation

We assume that the apprentice is given a *finite-horizon* MDP, which we call the modeling MDP, consisting of a finite set of states $\mathcal{S}$, a finite set of actions $\mathcal{A}$, a horizon $H$, and a reward function $R : \mathcal{S} \rightarrow \mathbb{R}$. We chose a finite horizon because our applications of interest are all episodic tasks. We also assume that we know the initial state distribution[1] $\mathbf{p^0} = \left( p_s^0 \right)_s$ and the transition probabilities $\boldsymbol{\theta} = (\theta_{sas'}^t)_{tsas'}$, where $\theta_{sas'}^t$ is the probability that the environment transitions from state $s$ to state $s'$ under action $a$ at time $t$ (this assumption can be relaxed; see Section 4.1). It is important to note that it is *not* the apprentice's objective to compute an optimal policy for the modeling MDP. Rather, the goal is to estimate the mentor's policy, and the modeling MDP is used to encode the apprentice's prior beliefs about that policy.

We further assume that we are given a data set $\mathcal{D}$ of state-action trajectories of the mentor acting in this environment. Concretely, $\mathcal{D} = \{x^i\}_{i=1}^m$, where $x^i$ is a sequence of $H$ state-action pairs; i.e., $x^i = (s_0^i, a_0^i), \ldots, (s_H^i, a_H^i)$. Our objective is to estimate the policy $\boldsymbol{\pi} = (\pi_{sa}^t)_{tsa}$ that governs the mentor's behavior, where $\pi_{sa}^t$ is the probability the mentor takes action $a$ in state $s$ at time $t$. The MAP estimate for the mentor's policy is given by

$$\hat{\boldsymbol{\pi}} = \arg \max_{\boldsymbol{\pi}} \log P(\mathcal{D} \mid \boldsymbol{\pi}) + \log P(\boldsymbol{\pi})$$

$$= \arg \max_{\boldsymbol{\pi}} \sum_{s,a,t} K_{sat} \log \pi_{sa}^t + \log P(\boldsymbol{\pi}),$$

---

[1]The notation $\mathbf{x} = (x_{ij})_{ij}$ denotes a vector $\mathbf{x}$ whose components are indexed by $i$ and $j$.

where $K_{sat}$ is the number of times in $\mathcal{D}$ that action $a$ is taken in state $s$ at time $t$. If the prior distribution $P(\boldsymbol{\pi})$ is uniform, then $\hat{\boldsymbol{\pi}}$ can be calculated analytically; the solution is just $\hat{\pi}_{sa}^t = \dfrac{K_{sat}}{\sum_a K_{sat}}$.

In this paper, we show how to assert a prior distribution $P(\boldsymbol{\pi})$ that gives greater weight to policies that have greater value in the MDP. Define the *value* of $\boldsymbol{\pi}$ to be

$$V(\boldsymbol{\pi}) = E\left[\sum_{t=0}^H R(s_t) \,\Big|\, \boldsymbol{\pi}, \boldsymbol{\theta}, s_0 \sim \mathbf{p^0}\right].$$

If we let $P(\boldsymbol{\pi}) = \exp(\alpha V(\boldsymbol{\pi}))$, then the MAP estimate is now given by

$$\hat{\boldsymbol{\pi}} = \arg\max_{\boldsymbol{\pi}} \sum_{s,a,t} K_{sat} \log \pi_{sa}^t + \alpha V(\boldsymbol{\pi}) \quad (1)$$
$$\triangleq \arg\max_{\boldsymbol{\pi}} L(\boldsymbol{\pi}).$$

Here, $\alpha$ can be viewed as a trade-off parameter that determines how much relative weight $P(\boldsymbol{\pi})$ assigns to high-value policies. Also note that $P(\boldsymbol{\pi})$ in this case is an unnormalized prior, as it does not necessarily intergrate to 1, and so (1) is perhaps more appropriately termed the estimate which maximizes a *penalized likelihood*.

In Section 4 we show how to efficiently find a $\hat{\boldsymbol{\pi}}$ that is provably a stationary point of $L(\boldsymbol{\pi})$.

## 4   Algorithm and Analysis

In this section, we present an outline of an iterative algorithm that converges to a stationary point $L(\boldsymbol{\pi})$, the function in Equation (1). In Section 4.2 we provide a detailed description of each iteration of the algorithm, and in Section 4.3 we sketch a proof of its convergence.

The trouble with finding the maximum of $L(\boldsymbol{\pi})$ directly is that the expression for $V(\boldsymbol{\pi})$, when expanded naively, contains $N^H$ terms. We can express $V(\boldsymbol{\pi})$ more compactly by using Bellman's equations, which yields the following optimization problem:

$$\max_{\boldsymbol{\pi}, \mathbf{V}} \quad \sum_{s,a,t} K_{sat} \log \pi_{sa}^t + \alpha \sum_s p_s^0 V_s^0$$

subject to:

$$\forall s, \,\forall\, t < H \qquad V_s^t = R(s) + \gamma \sum_{a,s'} \pi_{sa}^t \theta_{sas'}^t V_{s'}^{t+1} \quad (2)$$

$$\forall s \qquad V_s^H = R(s)$$
$$\forall s, t \qquad \sum_a \pi_{sa}^t = 1$$
$$\forall s, a, t \qquad \pi_{sa}^t \geq 0$$

where $\mathbf{V} = (V_s^t)_{ts}$ and $V_s^t$ is the value of the policy in state $s$ at time $t$. This problem is still difficult, however, since it involves nonconvex constraints — note that Bellman's equations (2) are bilinear in $\boldsymbol{\pi}$ and $\mathbf{V}$. To circumvent this, we will perform an *alternating maximization* instead. Let $\boldsymbol{\pi}^\tau = (\pi_{sa}^\tau)_{sa}$ and $\mathbf{V}^\tau = (V_s^\tau)_s$. In other words, $\boldsymbol{\pi} = (\boldsymbol{\pi}^0, \dots, \boldsymbol{\pi}^H)$ and $\mathbf{V} = (\mathbf{V}^0, \dots, \mathbf{V}^H)$. We will maximize $L(\boldsymbol{\pi})$ over just $\boldsymbol{\pi}^0$, then $\boldsymbol{\pi}^1$, and so on until $\boldsymbol{\pi}^H$, and then repeat the cycle until convergence (see Algorithm 1). In the iteration for $\boldsymbol{\pi}^\tau$, the values for $\boldsymbol{\pi}^0, \dots, \boldsymbol{\pi}^{\tau-1}, \boldsymbol{\pi}^{\tau+1}, \dots, \boldsymbol{\pi}^H$ are carried over from previous iterations and are held fixed while $\boldsymbol{\pi}^\tau$ is optimized. Taking this alternating approach has the effect of linearizing the constraints in (2), since $\mathbf{V}^{\tau+1}, \mathbf{V}^{\tau+2}, \dots, \mathbf{V}^H$ are not affected by changes to $\boldsymbol{\pi}^\tau$, and therefore can also be held fixed without impacting the maximization over $\boldsymbol{\pi}^\tau$.

Due to the linearization of the constraints in (2), each iteration of Algorithm 1 is just a convex optimization problem, and hence can be solved by any of a number of standard techniques, such as interior point methods. However, general-purpose methods are quite complex; fortunately they turn out to be unnecessary in this case. In Section 4.2, we describe a relatively simple procedure that solves this particular optimization problem in $O(|\mathcal{S}|^2 |\mathcal{A}| H + |\mathcal{S}||\mathcal{A}|(\log |\mathcal{A}| + \log |\mathcal{D}|))$ time.

---

**Algorithm 1** Find a stationary point of the log posterior.

---

Let $\boldsymbol{\pi}^\tau = (\pi_{sa}^\tau)_{sa}$ and $\boldsymbol{\pi} = (\boldsymbol{\pi}^0, \dots, \boldsymbol{\pi}^H)$.
Let $L(\boldsymbol{\pi}) = \sum_{s,a,t} K_{sat} \log \pi_{sa}^t + \alpha V(\boldsymbol{\pi})$.
Initialize $\tilde{\boldsymbol{\pi}}$ at random.
$\tau \leftarrow 0$.
**repeat**
    $\boldsymbol{\pi} \leftarrow \tilde{\boldsymbol{\pi}}$
    $\tilde{\boldsymbol{\pi}}^\tau = \arg\max_{\boldsymbol{\pi}^\tau} L(\boldsymbol{\pi})$
    $\tilde{\boldsymbol{\pi}} = (\boldsymbol{\pi}^0, \dots, \boldsymbol{\pi}^{\tau-1}, \tilde{\boldsymbol{\pi}}^\tau, \boldsymbol{\pi}^{\tau+1}, \dots, \boldsymbol{\pi}^H)$
    **if** $\tau = H$ **then**
        $\tau \leftarrow 0$
    **else**
        $\tau \leftarrow \tau + 1$
    **end if**
**until** $|L(\tilde{\boldsymbol{\pi}}) - L(\boldsymbol{\pi})|$ is as small as desired

---

### 4.1   When transition probabilities are unknown

So far, we have assumed that the transition probabilities $\boldsymbol{\theta}$ of the modeling MDP are given. Removing this assumption presents no special difficulty, since it is possible for our algorithm to jointly estimate $\boldsymbol{\theta}$ and $\boldsymbol{\pi}$ within the framework already presented. The idea will be to define new state and action spaces $\tilde{\mathcal{S}}$ and $\tilde{\mathcal{A}}$, and a new set of transition probabilities $\tilde{\boldsymbol{\theta}}$, in such a way that each parameter in the new set of unknowns $\tilde{\boldsymbol{\pi}}$ corresponds either to a parameter in $\boldsymbol{\pi}$ or a parameter in $\boldsymbol{\theta}$. Essentially, we fold the transition probabilities into the policy, and then replace them with a set of "dummy" transition probabilities. This reduction allows us to assume without loss of generality in our algorithm that $\boldsymbol{\theta}$ is known, and that everything unknown about the MDP is embodied in the policy $\boldsymbol{\pi}$.

Concretely, let $\tilde{\mathcal{S}} = \mathcal{S} \cup (\mathcal{S} \times \mathcal{A})$ and $\tilde{\mathcal{A}} = \mathcal{A} \cup \mathcal{S}$. We

define $\tilde{\boldsymbol{\theta}}$ as

$$\tilde{\theta}_{\tilde{s}\tilde{a}\tilde{s}'}^t = \begin{cases} 1 & \text{if } \tilde{s} \in \mathcal{S}, \ \tilde{a} \in \mathcal{A}, \ \text{and } \tilde{s}' = (\tilde{s}, \tilde{a}); \ \text{or} \\ & \text{if } \tilde{s} \in (\mathcal{S} \times \mathcal{A}), \ \tilde{a} \in \mathcal{S}, \ \text{and } \tilde{s}' = \tilde{a} \\ 0 & \text{otherwise.} \end{cases}$$

Put differently, when we are in state $\tilde{s} = s$ and take action $\tilde{a} = a$, the environment deterministically transitions to "state" $\tilde{s}' = (s, a)$. And when we are in "state" $\tilde{s} = (s, a)$ and take "action" $\tilde{a} = s'$, the environment deterministically transitions to state $\tilde{s}' = s'$.

One last modification is needed: we define a new state $\tilde{s}^*$, with $R(\tilde{s}^*) = -\infty$, and set $\tilde{\theta}_{\tilde{s}\tilde{a}\tilde{s}^*}^t = 1$ whenever $\tilde{s}$ and $\tilde{a}$ do not make sense together, i.e., when $\tilde{s} \in \mathcal{S}$ and $\tilde{a} \in \mathcal{S}$, or when $\tilde{s} \in (\mathcal{S} \times \mathcal{A})$ and $\tilde{a} \in \mathcal{A}$. This will force $\tilde{\pi}_{\tilde{s}\tilde{a}}^t = 0$ in these cases.

So we have the following equivalences between the old and new parameters:

$$\begin{array}{lll} \tilde{\pi}_{\tilde{s}\tilde{a}}^t & \Leftrightarrow & \pi_{sa}^t \quad \text{if } \tilde{s} = s \text{ and } \tilde{a} = a \\ \tilde{\pi}_{\tilde{s}\tilde{a}}^t & \Leftrightarrow & \theta_{sas'}^t \quad \text{if } \tilde{s} = (s, a) \text{ and } \tilde{a} = s' \end{array}$$

Note that, when applying this reduction, the prior $P(\tilde{\boldsymbol{\pi}}) = P(\boldsymbol{\pi}, \boldsymbol{\theta})$ assigns greater weight to policies and transition probabilities that *jointly* have high value.

### 4.2 Optimization procedure

Recall that $\mathbf{V}^\tau = (V_s^\tau)_s$, $\boldsymbol{\pi}^\tau = (\boldsymbol{\pi}_{sa}^\tau)_{sa}$, and $\boldsymbol{\pi} = (\boldsymbol{\pi}^0, \dots, \boldsymbol{\pi}^H)$. In each iteration of Algorithm 1, we maximize $L(\boldsymbol{\pi})$ over $\boldsymbol{\pi}^\tau$, for some $\tau \in \{0, \dots, H\}$. When $\tau \neq H$, the corresponding convex optimization (after dropping constant terms) is[2]

$$\max_{\boldsymbol{\pi}^\tau, \mathbf{V}^0, \dots, \mathbf{V}^\tau} \sum_{s,a} K_{sa\tau} \log \pi_{sa}^\tau + \alpha \sum_s p_s^0 V_s^0$$

subject to:

$$\forall s, \ \forall\, t \leq \tau \qquad V_s^t = R(s) + \gamma \sum_{a,s'} \pi_{sa}^t \theta_{sas'}^t V_{s'}^{t+1}$$

$$\forall s \qquad \sum_a \pi_{sa}^\tau = 1$$

$$\forall s, a \qquad \pi_{sa}^\tau \geq 0.$$

Recall that $\boldsymbol{\pi}^0, \dots, \boldsymbol{\pi}^{\tau-1}, \boldsymbol{\pi}^{\tau+1}, \dots, \boldsymbol{\pi}^H$ and $\mathbf{V}^{\tau+1}, \dots, \mathbf{V}^H$ are constants in this problem; their values are carried over from previous iterations.

To solve the optimization, we need to find a solution to the *KKT conditions*, i.e., a solution $(\boldsymbol{\pi}^\tau, \mathbf{V}^0, \dots, \mathbf{V}^\tau, \boldsymbol{\lambda})$ that is both feasible and also satifies

$$\nabla \mathcal{L}(\boldsymbol{\pi}^\tau, \mathbf{V}^0, \dots, \mathbf{V}^\tau, \boldsymbol{\lambda}) = 0$$
$$\forall s, a \qquad \lambda_{sa}^\pi \geq 0$$
$$\forall s, a \qquad \lambda_{sa}^\pi \cdot \pi_{sa}^\tau = 0$$

------

[2] The solution for the $\tau = H$ case is similar to the procedure described in this section, except it is even simpler, so we omit its discussion.

where $\boldsymbol{\lambda} = \{\lambda_{st}^V, \lambda_s^\pi, \lambda_{sa}^\pi \mid s \in \mathcal{S}, a \in \mathcal{A}, t \leq \tau\}$, the Lagrangian $\mathcal{L}(\boldsymbol{\pi}^\tau, \mathbf{V}^0, \dots, \mathbf{V}^\tau, \boldsymbol{\lambda})$ is given by

$$\mathcal{L}(\boldsymbol{\pi}^\tau, \mathbf{V}^0, \dots, \mathbf{V}^\tau, \boldsymbol{\lambda}) =$$
$$\sum_{s,a} K_{sa\tau} \log \pi_{sa}^\tau + \alpha \sum_s p_s^0 V^0 +$$
$$\sum_{\substack{s \\ t \leq \tau}} \lambda_{st}^V \left[ R_s + \gamma \sum_{a,s'} \pi_{sa}^t \theta_{sas'}^t V_{s'}^{t+1} - V_s^t \right] +$$
$$\sum_s \lambda_s^\pi \left[ 1 - \sum_a \pi_{sa}^\tau \right] +$$
$$\sum_{s,a} \lambda_{sa}^\pi \cdot \pi_{sa}^\tau$$

and the gradient of $\mathcal{L}$ is taken with respect to $(\boldsymbol{\pi}^\tau, \mathbf{V}^0, \dots, \mathbf{V}^\tau)$.

Below we outline a three-step procedure for finding $(\boldsymbol{\pi}^\tau, \mathbf{V}^0, \dots, \mathbf{V}^\tau, \boldsymbol{\lambda})$ that satisfies the KKT conditions.

#### 4.2.1 Step 1: Find the $\lambda_{st}^V$'s

From the KKT conditions, we must have that

$$\frac{\partial \mathcal{L}}{\partial V_s^t} = 0 \quad \forall s, \ \forall\, t \leq \tau.$$

This yields

$$\begin{aligned} \lambda_{s0}^V &= \alpha p_s^0 \\ \lambda_{st}^V &= \gamma \sum_{s',a} \lambda_{s't-1}^V \pi_{s'a}^{t-1} \theta_{s'as}^t \ \text{ for } 0 < t \leq \tau \end{aligned}$$

which allows us to inductively compute all the $\lambda_{st}^V$'s. We can see from this expression that

$$\lambda_{st}^V = \alpha \gamma^t \Pr[s_t = s \mid \boldsymbol{\pi}],$$

i.e., $\lambda_{st}^V$ is equal to the occupancy probability of state $s$ at time $t$ under policy $\boldsymbol{\pi}$, but scaled by $\alpha \gamma^t$.

#### 4.2.2 Step 2: Find the $\lambda_s^\pi$'s, $\lambda_{sa}^\pi$'s and $\pi_{sa}^\tau$'s

To simplify notation, define

$$\begin{aligned} B_{sa\tau} &\triangleq \gamma \lambda_{s\tau}^V \sum_{s'} \theta_{sas'}^\tau V_{s'}^{\tau+1} \\ \mathcal{A}_s^0 &\triangleq \{a \in \mathcal{A} \mid K_{sa\tau} = 0\} \\ \mathcal{A}_s^{\neg 0} &\triangleq \mathcal{A} \setminus \mathcal{A}_s^0. \end{aligned}$$

Let us focus on a particular state $s$. We know that $\sum_a \pi_{sa}^\tau = 0$ and $\pi_{sa}^\tau \geq 0$ for all $a$. Suppose we can find a value of $\lambda_s^\pi$ such that

$$\sum_{a \in \mathcal{A}_s^{\neg 0}} \frac{K_{sa\tau}}{\lambda_s^\pi - B_{sa\tau}} = 1 \qquad (3)$$

$$\frac{K_{sa\tau}}{\lambda_s^\pi - B_{sa\tau}} \geq 0 \quad \forall a \in \mathcal{A}_s^{\neg 0}. \qquad (4)$$

If it happens that $\lambda_s^\pi \geq \max_{a \in \mathcal{A}} B_{sa\tau}$, then we can satisfy all the relevant KKT conditions by setting

$$
\begin{aligned}
\lambda_{sa}^\pi &= 0 & \forall a \in \mathcal{A}_s^{-0} \\
\lambda_{sa}^\pi &= \lambda_s^\pi - B_{sa\tau} & \forall a \in \mathcal{A}_s^0 \\
\pi_{sa}^\tau &= \frac{K_{sa\tau}}{\lambda_s^\pi - B_{sa\tau}} & \forall a \in \mathcal{A}_s^{-0} \\
\pi_{sa}^\tau &= 0 & \forall a \in \mathcal{A}_s^0.
\end{aligned}
$$

On the other hand, if $\lambda_s^\pi < \max_{a \in \mathcal{A}} B_{sa\tau}$ for the value of $\lambda_s^\pi$ that solves (3) and (4), then we can satisfy the relevant KKT conditions by first letting $\lambda_s^\pi = \max_{a \in \mathcal{A}} B_{sa\tau}$, and then setting

$$
\begin{aligned}
\lambda_{sa}^\pi &= 0 & \forall a \in \mathcal{A}_s^{-0} \\
\lambda_{sa}^\pi &= \lambda_s^\pi - B_{sa\tau} & \forall a \in \mathcal{A}_s^0 \\
\pi_{sa}^\tau &= \frac{K_{sa\tau}}{\lambda_s^\pi - B_{sa\tau}} & \forall a \in \mathcal{A}_s^{-0} \\
\pi_{sa}^\tau &= 0 & \forall a \in \mathcal{A}_s^0 \setminus \{a^*\} \\
\pi_{sa^*}^\tau &= 1 - \sum_{a \in \mathcal{A}_s^{-0}} \pi_{sa}^\tau.
\end{aligned}
$$

where $a^* = \arg \max_{a \in \mathcal{A}} B_{sa\tau}$.

So it remains to show that we can easily find a $\lambda_s^\pi$ that solves (3) and (4). Define

$$
\begin{aligned}
B_{\max} &\triangleq \max_{a \in \mathcal{A}_s^{-0}} B_{sa\tau} \\
K_{\max} &\triangleq \max_{a \in \mathcal{A}_s^{-0}} K_{sa\tau} \\
K_{\min} &\triangleq \min_{a \in \mathcal{A}_s^{-0}} K_{sa\tau}
\end{aligned}
$$

and observe that

$$
\begin{aligned}
\lambda_s^\pi &= K_{\min} + B_{\max} \\
\Rightarrow \sum_a \frac{K_{sa\tau}}{\lambda_s^\pi - B_{sa\tau}} &\geq 1
\end{aligned}
$$

and

$$
\begin{aligned}
\lambda_s^\pi &= |\mathcal{A}| \cdot K_{\max} + B_{\max} \\
\Rightarrow \sum_a \frac{K_{sa\tau}}{\lambda_s^\pi - B_{sa\tau}} &\leq 1.
\end{aligned}
$$

Moreover, the left-hand side of (3) is strictly monotone in $\lambda_s^\pi$, and $\lambda_s^\pi \in [K_{\min} + B_{\max}, |\mathcal{A}| \cdot K_{\max} + B_{\max}]$ satisfies (4). Putting all this together with the Intermediate Value Theorem, we conclude that there exists a unique $\lambda_s^\pi \in [K_{\min} + B_{\max}, |\mathcal{A}| \cdot K_{\max} + B_{\max}]$ that satisfies (3) and (4), so we can use a simple root-finding algorithm such as the bisection method to approximate it within a constant $\epsilon$.

#### 4.2.3   Step 3: Find the $V_s^t$'s

Since we know the $\pi_{sa}^\tau$'s now, all the $V_s^0, \ldots, V_s^\tau$'s can be computed inductively.

$$
V_s^t = R(s) + \gamma \sum_{a,s'} \pi_{sa}^t \theta_{sas'}^t V_{s'}^{t+1} \quad \forall s \in \mathcal{S}, \, \forall t \leq \tau.
$$

#### 4.2.4   Running time

Recall that $\mathcal{S}$ and $\mathcal{A}$ are state and action spaces, respectively, $\mathcal{D}$ is the data set of state-action trajectories, $H$ is the length of the horizon, and $\epsilon$ is the approximation error of the root-finding algorithm in Step 2.

Steps 1 and 3 both take $O(|\mathcal{S}|^2|\mathcal{A}|H)$ time, and step 2 takes $O(|\mathcal{S}||\mathcal{A}|(\log|\mathcal{A}| + \log|\mathcal{D}| + \log\frac{1}{\epsilon}))$ time (the log factors are from the root-finding algorithm, e.g. the bisection method, for which the running time is logarithmic in the size of the interval being searched). This yields a total running time of $O(|\mathcal{S}|^2|\mathcal{A}|H + |\mathcal{S}||\mathcal{A}|(\log|\mathcal{A}| + \log|\mathcal{D}| + \log\frac{1}{\epsilon}))$ for each iteration of Algorithm 1. In practice, we have observed that only a handful of iterations are required for convergence. By comparison, determining the optimal policy takes $O(|\mathcal{S}|^2|\mathcal{A}|H)$ time.

### 4.3   Analysis

In this section, we sketch a proof that the sequence of estimates produced by Algorithm 1 converges to a limit that is a stationary point of $\bar{L}(\pi)$, the function in Equation (1). This guarantee is similar to the one typically cited for the EM algorithm [2]; in fact, the convergence theorem used in the proof sketch below is the same tool used by Wu [13] in his analysis of EM. A complete proof of Theorem 1 is available in the supplement for this paper [11].

**Theorem 4.1.** *Algorithm 1 converges to a stationary point of $L(\pi)$.*

*Proof sketch.* Let $\Omega$ be the set of all policies. We will need to assume that each maximization in Algorithm 1 finds a point in the interior of $\Omega$ (a similar assumption is made in Wu's proof of the convergence of the EM algorithm [13]). We can view Algorithm 1 as defining $H$ distinct point-to-set maps $\{M_\tau\}_{\tau=0}^H$ on $\Omega$, each corresponding to an optimization over a different $\pi^\tau$. In other words, $\tilde{\pi} \in M_\tau(\pi)$ if $\tilde{\pi}$ is a solution to the problem of maximizing $L(\pi)$ over just the variables in $\pi^\tau$ (recall that $\pi = (\pi^0, \ldots, \pi^H)$). Let $M^A = M_H \circ M_{H-1} \cdots \circ M_0$, i.e., $M^A$ is the point-to-set map defined by one complete cycle of optimizations.

By Convergence Theorem A of Zangwill [14], Algorithm 1 converges to a stationary point of $L$ if: (a) $\Omega$ is compact, (b) for all $\tilde{\pi} \in M^A(\pi)$, $L(\tilde{\pi}) \geq L(\pi)$, (c) whenever $\pi$ is not a stationary point of $L$, then for all $\tilde{\pi} \in M^A(\pi)$, we have $L(\tilde{\pi}) > L(\pi)$, and (d) $M^A$ is a closed map.

Conditions (a), (b) and (c) are fairly straightforward to establish. The last condition (d) is more difficult, but this can be proved by observing that $L$ is continuous, and then applying Proposition 7 and Theorem 8 of Hogan [5]. $\quad\square$

## 5   Experiments

Using synthetic environments, we compared the value-based prior to two similar algorithms proposed by other authors. We also investigated our algorithm's sensitivity to the value of the mentor's policy. We review the other methods below, the synthetic environments in Section 5.1, and our experiments in Sections 5.2 and 5.3. Additional experiments are presented in the supplement for this paper [11].

Recall that Price *et al* [7] proposed to model the mentor's policy using a Dirichlet distribution. In their scheme, the policy at each state is assigned a prior distribution $P_s(a; \boldsymbol{\beta}) = Dir(\boldsymbol{\beta})$, where $\boldsymbol{\beta}$ is a $|\mathcal{A}|$-length vector of positive reals. Let $\mathcal{A}_s^o$ be the set of optimal actions at state $s$. We define each $P_s(a; \boldsymbol{\beta})$ so that $\beta_a = \dfrac{\alpha}{|\mathcal{A}_s^o|}$. This amounts to asserting a prior belief that the mentor's policy is an optimal policy. Note that $\alpha$ plays a similar role here as it does in Equation (1), in that it reflects the degree to which the prior is concentrated on high-value policies.

Similarly, recall the temporal-difference-like algorithm developed by Henderson *et al* [4], in which the usual $Q$ values are modified so that the optimal policy is more similar to the policy that generated the data. Although it is difficult to describe succinctly, their algorithm employs a tunable parameter $\alpha$, which controls the trade-off between optimality and imitation, just as it does in our algorithm. Since TD techniques do not assume that transition probabilities are given, we use the reduction described in Section 4.1 when comparing with our method.

## 5.1 Maze environments

We used maze environments for all of our experiments. Each maze was a 30-by-30 grid, with the start state in one corner and the goal state, containing a large positive reward, in the opposite corner. Movement in a maze was in the four compass directions, but taking a move action risked a 30% chance of landing in a random adjacent cell. Also, obstacles (negative rewards) were randomly placed in 15% of the cells in each maze, with each having a magnitude that was, on average, 2/3 as large as the goal state's positive reward. Finally, the time horizon was set to 90, which was sufficient to allow even meandering policies to eventually reach the goal state.

Our environments had one additional feature that was introduced to make the comparison between the various algorithms more interesting. We found that the optimal action in each state typically had substantially larger value than any other action. So a prior that assigned greatest weight to the highest value policies essentially assigned greatest weight to a *single* policy, i.e., the policy that takes the optimal action in every state. In such circumstances, we did not expect to observe an advantage to using a value-based prior over a Dirichlet prior. To simulate a scenario where there are many diverse high-value policies, we introduced a "twin" action for every original action, i.e., a separate action that has exactly the same effect on the environment.

## 5.2 Comparison to other methods

### 5.2.1 Experimental setup

For each maze environment, we generated data sets of state-action trajectories from an optimal policy for the maze.[3] However, when estimating that policy from data, we supplied each algorithm with just the location and size of the goal reward, and *not* the locations or sizes of the obstacles. Effectively, each algorithm assigned the highest prior probability to a policy that moved directly towards the goal, ignoring obstacles altogether. So, from the perspective of each algorithm, the mentor's policy had high value, but was suboptimal.

### 5.2.2 Results

Figure 1 compares the value-based prior to the Dirichlet prior suggested by Price *et al* [7]. First, note that our algorithm is much more robust to the value of the trade-off parameter $\alpha$; we varied $\alpha$ over three orders of magnitude, and the value-based prior improved the accuracy of estimated policy throughout that range. This is important, as we are not proposing a principled way to set the value of $\alpha$, except to point out that it should generally increase with the value of the mentor's policy. Second, although the Dirichlet prior provided a more accurate estimate for smaller data sets for certain values of $\alpha$, that advantage soon became a disadvantage as the amount of data was increased. To understand why, recall that in our maze environment, there are many diverse policies that each have high value. The value-based prior assigns the same weight to every policy that has the same value, even if the policies themselves are quite different. But a Dirichlet prior is forced to encode the belief that a *particular* policy is most probable. If this policy differs from the mentor's policy, then it will skew the estimation, even if both are high value polices.

Figure 2 compares the value-based prior to the hybrid reinforcement/supervised learning algorithm proposed by Henderson *et al* [4]. For the value-based prior, the reduction described in Section 4.1 was applied, since the hybrid algorithm does not assume that the transition probabilities given. Note that the value-based prior initially provides an inferior estimate than the naive method that uses no prior; we suspect this is because the algorithm at that stage is using poor approximations of the transition probabilities to compute value function in the modeling MDP. Nevertheless, as the number of samples increases, the value-based prior eventually provides an advantage. The performance of the hybrid algorithm is perhaps not indicative of its general usefulness, as it may not have been designed with this particular application in mind.

## 5.3 Sensitivity to policy value

We also investigated how sensitive our algorithm is to the value of the mentor's policy.

---

[3]Since there were always at least two optimal actions in each state, per Section 5.1, we randomly chose one of them to always take.

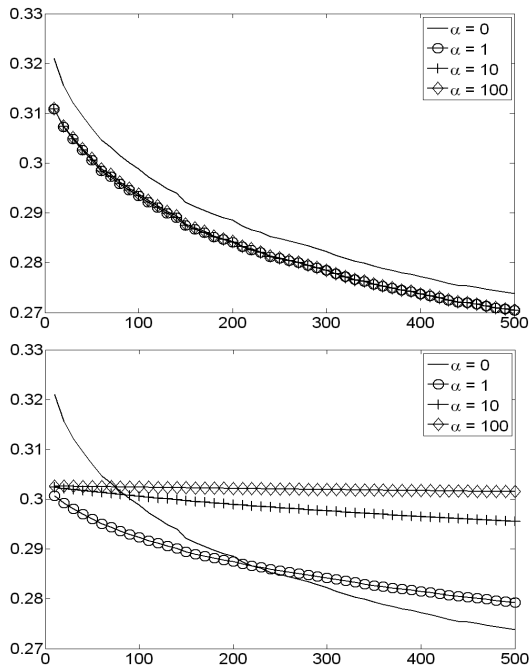

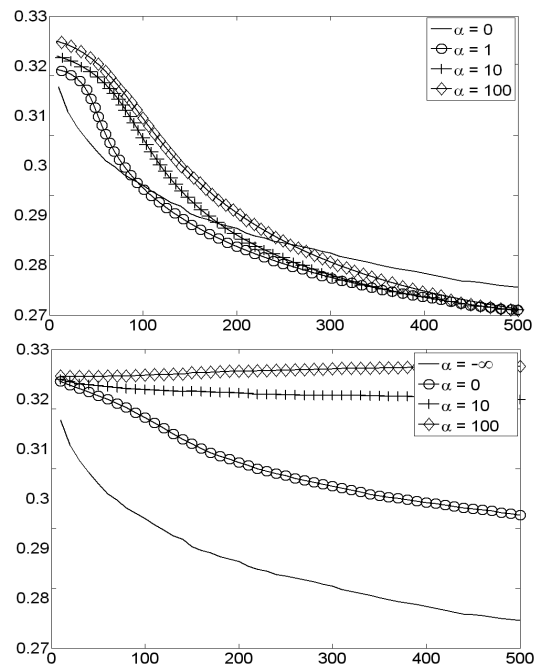

Figure 1: Top: Performance of the value-based prior. Bottom: Performance of the Dirichlet prior. The x-axis indicates the number of state-action trajectories in the data set, and the y-axis indicates the RMS error of the estimated policy with respect to the mentor's policy. Each line in each graph is the average estimation error for 50 mazes. $\alpha$ is a trade-off parameter; $\alpha = 0$ corresponds to not using any prior at all.

Figure 2: Top: Performance of the value-based prior. Bottom: Performance of the hybrid reinforcement/supervised learning algorithm. Details are the same as for Figure 1, except that for the value-based prior, the reduction described in Section 4.1 has been applied, and in the case of the hybrid algorithm, $\alpha = -\infty$ corresponds to ignoring rewards and simply imitating the behavior in the data.

### 5.3.1  Experimental setup

To create policies with a variety of values, we used the following procedure. In each maze environment, we computed an optimal policy $\pi^*$. We then randomly selected $\delta$ fraction of the states, and in each state swapped the optimal action in $\pi^*$ with a randomly chosen action. We also added a small Gaussian perturbation (mean 0.5, variance $\sigma^2$) to each state-action probability, and renormalized appropriately. By carefully varying $\delta$ and $\sigma^2$, we were able to produce policies whose values were distributed in a range of 70% to 100% of the optimal value.

### 5.3.2  Results

Figure 3 depicts the performance of our method for estimating policies with various values. As one might expect, performance degraded as the mentor's policy's value decreased. Nonetheless, we found that the value-based prior improves estimation even when the mentor's policy's value is reasonably far from optimal — as low as 80% of the optimal value.

## 6   Summary and Future Work

We have presented a novel approach to imitation learning, where an apprentice uses the value function of an MDP to assert a prior belief on a mentor's policy, and have provided both theoretical and empirical evidence that our algorithm is sound and useful. Our analysis suggests that our algorithm, similar to the EM algorithm, will often find a local maximum of the log posterior distribution $L(\pi)$ (Equation (1)). Our experiments indicate that a value-based prior is robust in at least two senses: it is effective over a wide range of values for the trade-off parameter $\alpha$, and it is effective even when the mentor's policy is suboptimal.

The value-based prior described here differs from the prior distributions used in previous approaches to imitation learning [3, 4, 7] in several significant ways. Unlike in earlier methods, the value-based prior was not chosen for the sake of mathematical convenience (it is in fact quite inconvenient to work with), but rather to allow an apprentice to assert a very natural belief about the mentor's behavior — that the mentor is reward-seeking. Additionally, unlike the Dirichlet prior, the value-based prior is not separable over states. In other words, evidence for the mentor's policy at one state can affect the maximum likelihood estimate of the mentor's policy at other states, a useful "coupling" property. Also, the value-based prior assigns high probability to all mentor policies that lead to high value, allowing the apprentice to remain agnostic about which particular distribution over actions the mentor takes in each state.

We are currently extending this work in several directions. First, we are developing an algorithm that can be applied to infinite horizon problems, a well-studied setting with many applications. Second, we are attempting to strengthen our convergence result to prove that Algorithm 1 will always find the global optimum of the posterior defined in Equation (1). Finally, we would like to introduce function approximation into this framework, so that we can apply our method to larger state spaces.

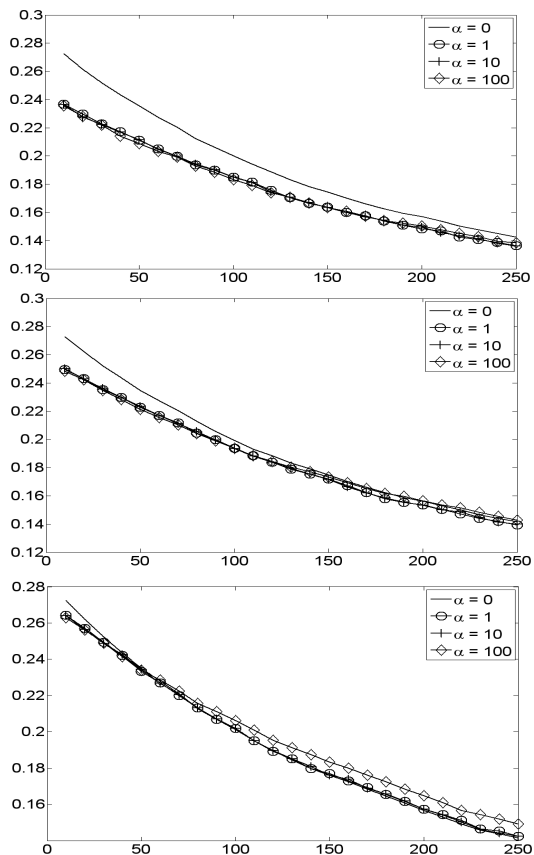

Figure 3: Performance of the value-based prior for policies with values approximately 70-90% of the optimal value. Axes and legend are the same as for Figure 1. Each line in each graph is the average estimation error for 50 policies (10 policies each from 5 maze environments). Top: Policies that have average value 89.6% of the optimal value, with std dev 1.3%. Middle: Policies that have average value 80.9% of the optimal value, with std dev 3.2%. Bottom: Policies that have average value 72.9% of the optimal value, with std dev 1.1%.

### Acknowledgements

The authors were supported by the NSF under grant IIS-0325500. This work benefited greatly from discussions with Esther Levin, Michael Littman, and Mazin Gilbert. We would also like to thank Srinivas Bangalore and Jason Williams for their suggestions for applications and other helpful comments. Finally, we thank the anonymous referees for their feedback.